\pdfoutput=1

\documentclass[11pt]{article}

\usepackage[final]{acl}

\usepackage{times}
\usepackage{latexsym}

\usepackage[T1]{fontenc}

\usepackage[utf8]{inputenc}

\usepackage{microtype}

\usepackage{url}
\usepackage{amsmath}
\usepackage{multirow}
\usepackage{booktabs}
\usepackage{amssymb}
\usepackage{mdframed}

\usepackage{inconsolata}

\usepackage{graphicx}

\title{ReachAgent: Enhancing Mobile Agent via Page Reaching and Page Operation}

\author{\bf Qinzhuo Wu, Wei Liu\thanks{\,\,\,\,Corresponding author.}, Jian Luan, Bin Wang \\
  XiaoMi AI Lab \\
  \texttt{\{wuqinzhuo, liuwei40, luanjian, wangbin11\}@xiaomi.com} \\}

\begin{document}
\maketitle
\begin{abstract}
Recently, mobile AI agents have gained increasing attention. Given a task, mobile AI agents can interact with mobile devices in multiple steps and finally form a GUI flow that solves the task. However, 
existing agents tend to focus on the most task-relevant elements at each step, leading to local optimal solutions and ignoring the overall GUI flow.
To address this issue, we constructed a training dataset called MobileReach, which breaks the task into page reaching and operation subtasks.
Furthermore, we propose ReachAgent, a two-stage framework that focuses on improving its task-completion abilities.
It utilizes the page reaching and page operation subtasks, along with reward-based preference GUI flows, to further enhance the agent.
Experimental results show that ReachAgent significantly improves the Intersection over Union (IoU) Accuracy and Text Accuracy by \textbf{7.12\% and  7.69\%} on step-level and \textbf{4.72\% and 4.63\%} on task-level compared to the SOTA agent. 
Our data and code will be released upon acceptance.

\end{abstract}

\section{Introduction}
With the quick advancement of visual language models (VLMs)  \cite{wang2023cogvlm,bai2023qwen}, it has become more feasible to create mobile AI agents that can operate mobile devices \cite{yang2023appagent, ding2024mobileagent, li2020mapping}. 
Some early works \cite{yang2023appagent,zhang2024android,yan2023gpt} have attempted to combine powerful general VLM models, such as GPT-4V \cite{openai2023gpt4}, with prompt engineering and retrieval modules to generate UI actions for mobile control. 
Other works have used GUI navigation datasets \cite{rawles2023android,zhang2024android} to fine-tune the base VLMs to improve their stability in generating UI actions. Subsequent works \cite{baechler2024screenai, wu2024mobilevlm} have built mobile-specific datasets to pre-train VLMs so that the model can better understand graphical user interface (GUI) elements and page structure.
Specifically, mobile AI agents use the VLM as their base model and the mobile device as the environment. The agent iteratively interacts with the mobile's GUI for each mobile control task, by generating actions based on environmental information and executing those actions to update the GUI environment. This interactive process forms a GUI flow.

\begin{figure}[!t]
  \centering
  \includegraphics[width=1\columnwidth]{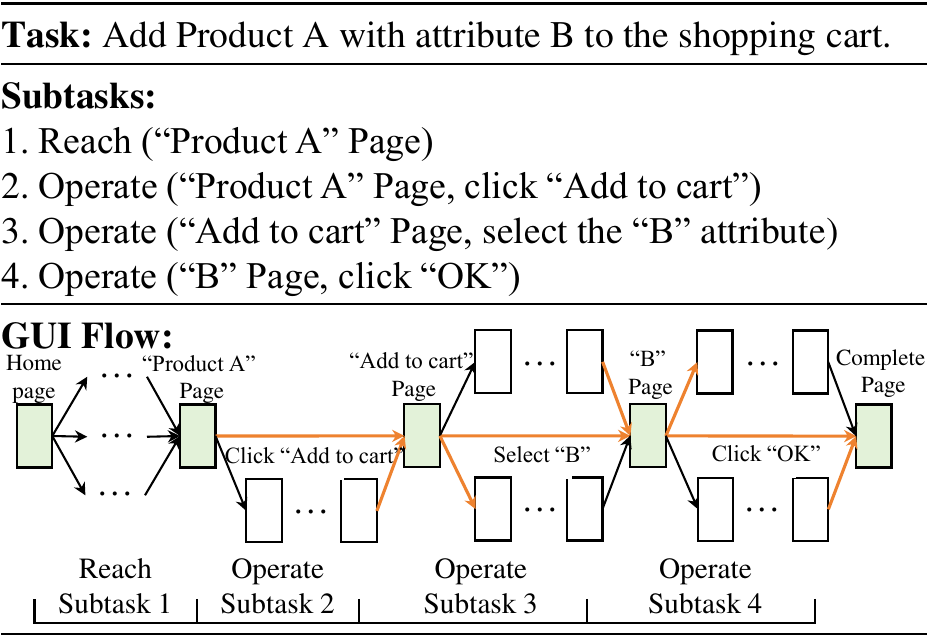}
  \caption{An example of a task and its subtasks and possible GUI flows. Green boxes represent the pages that need to be reached, and orange arrows represent the actions that need to be operated. To complete this task, the agent must reach 5 pages and do 3 operations.}
  \label{figure1-1}
\end{figure}

\begin{figure*}[!t]
  \centering
  \includegraphics[width=0.93\textwidth]{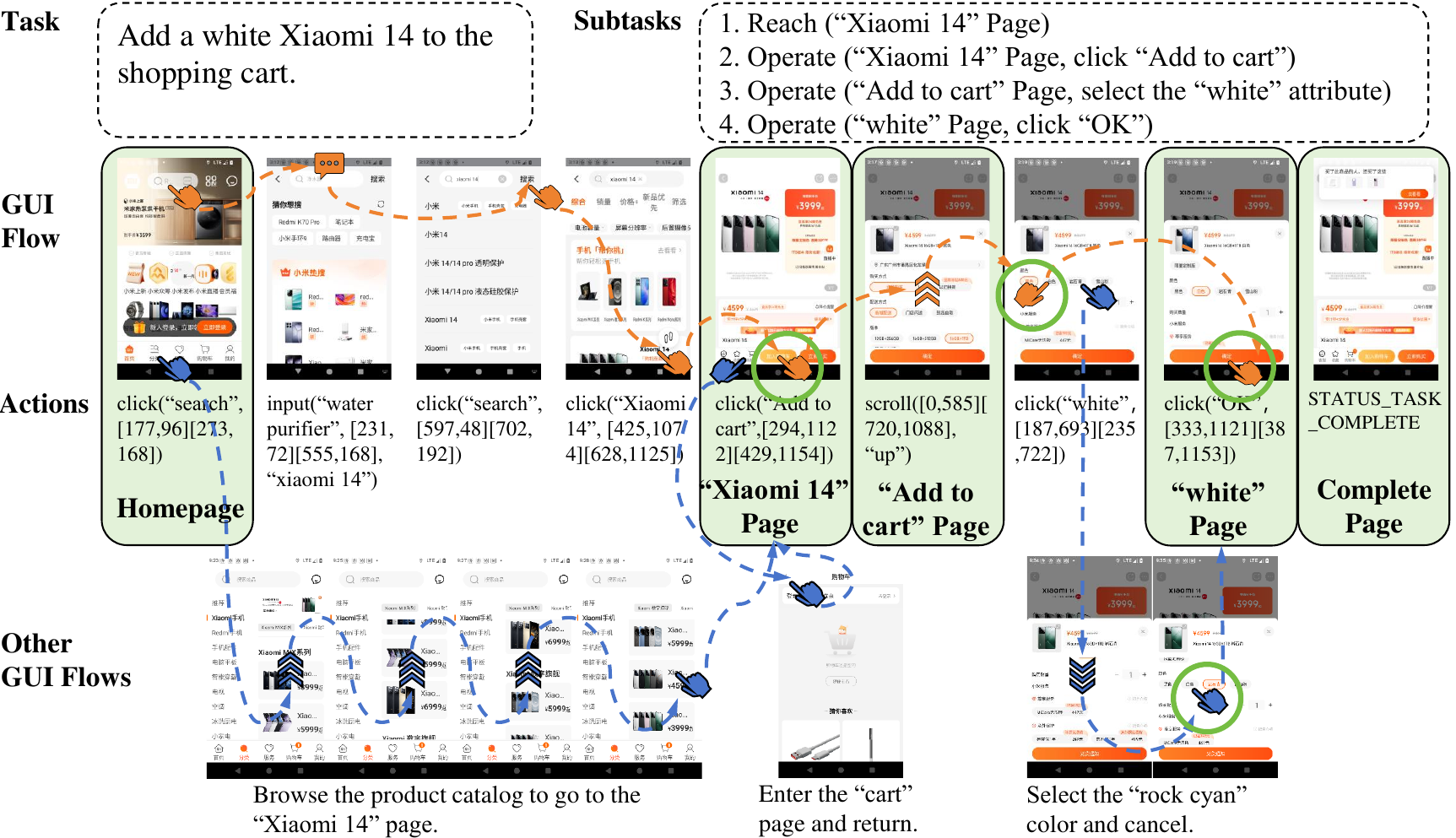}
  \caption{The complete 9-step GUI flow for a task. Green boxes represent the pages that need to be reached, and green circles represent the operations that need to be done. Orange arrows are the actions in the golden flow. Blue arrows are the actions in other GUI flows. Both the orange and blue flows can complete the task.}
  \label{figure1-2}
\end{figure*}

Although existing mobile AI agents have shown good performance, they still have some limitations. 
Existing agents focus on interacting with the most task-relevant elements of the current page, and ignore whether the entire GUI flow solves the task.
This method sometimes causes them to focus on single-step action accuracy and select tasks-related actions greedily, thus falling into a local optimal solution.
As shown in Figure \ref{figure1-2}, for the task of adding Xiaomi 14 to the shopping cart, the agent should first reach the product page, but the homepage does not contain elements related to the product name. In this case, it might go directly from the homepage to the shopping cart page.



Therefore, we break down the task into several subtasks and focus on the agent's subtask completion abilities. Intuitively, these subtasks can be divided into two categories, reach and operate. 
Reach only requires the agent to reach a specific page, regardless of the path taken. Operate requires the agent to reach a specific page and perform some specific operations.
Figure \ref{figure1-2} shows a task and its GUI flow, consisting of a 9-step chain, the key to solving this task is to reach the product page first, and then click the "Add to Cart" pop-up window and select the attributes and "OK" button in the pop-up window in sequence, as shown in Figure \ref{figure1-1}. For the product page reach subtask, it can be done either through product search or by browsing the product catalog. Similarly, the operate subtask only requires the agent to select the "B" attribute in the pop-up window, and browsing other attributes does not affect the task completion.


To address these limitations, we collected a training dataset called MobileReach, which contains three types of tasks: page navigation, page reaching, and page operation.
In addition, we propose ReachAgent, a two-stage framework that focuses on the ability to complete subtasks.
In the first stage, in addition to page navigation abilities, it also learns how to reach a specified page and complete operations on a specified page.
In the second stage, a 4-level reward function is used to weigh different GUI flows and construct preference data to further reinforce ReachAgent, 
Furthermore, an action alignment mechanism is proposed to reduce the difficulty of action generation.
The main contributions of this paper can be summarized as follows:

$\bullet$ We break down the mobile control task into page reaching and page operation subtasks, and construct a mobile control dataset called MobileReach consisting of three types of tasks.

$\bullet$ We proposed ReachAgent, a two-stage framework that utilizes page reaching and page operation to enhance the agent's subtask completion abilities.
In addition, it uses reinforcement learning to further enhance the agent's overall task completion ability.
ReachAgent includes an action alignment mechanism that decreases the number of candidate actions, thereby reducing the difficulty of the task.


$\bullet$ Experimental results show that ReachAgent outperforms existing state-of-the-art agents. It also demonstrates stronger page reaching and operation abilities.

\begin{figure*}[!t]
  \centering
  \includegraphics[width=0.9 \textwidth]{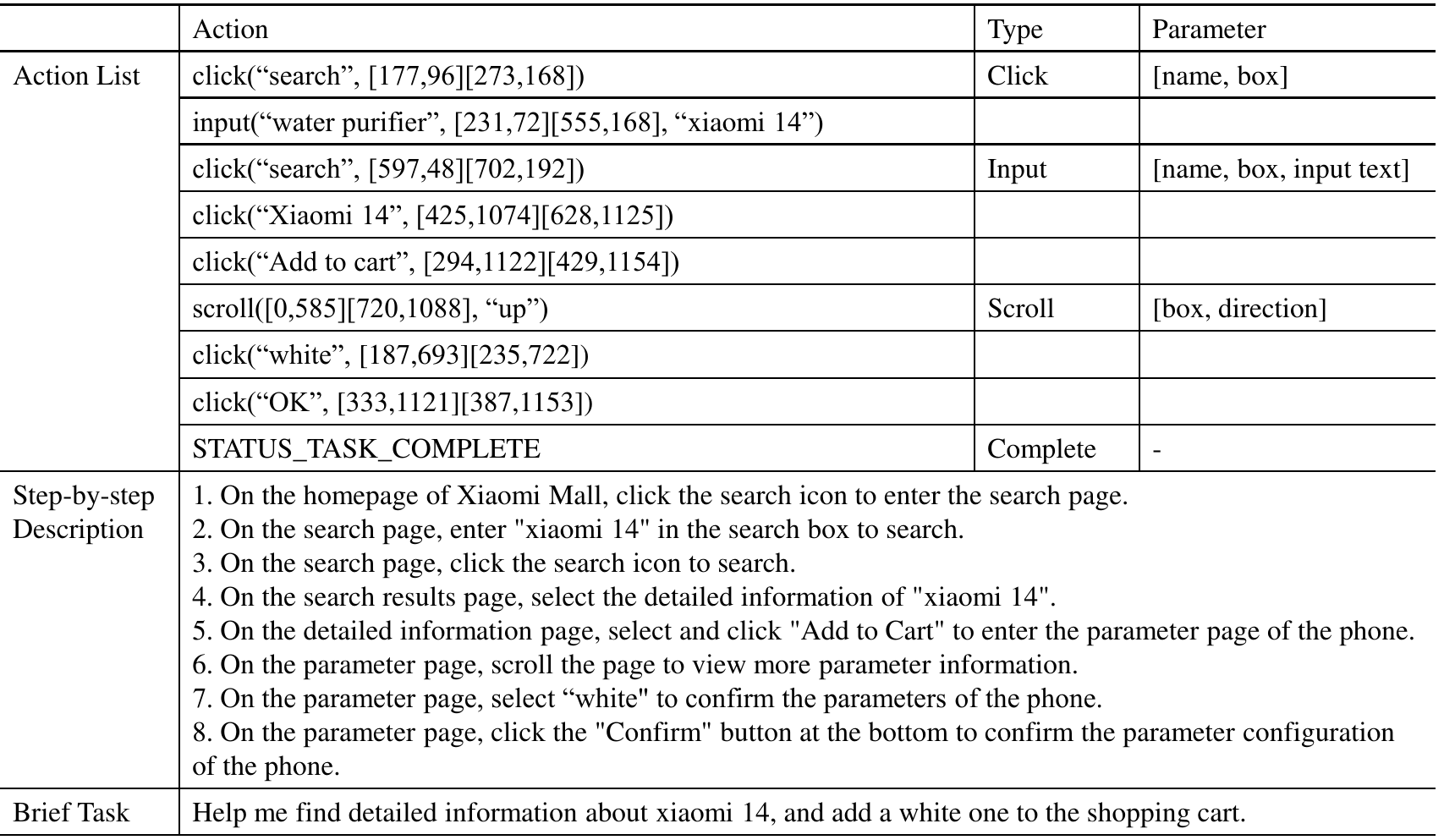}
  \caption{Actions and tasks for a GUI flow. The step-by-step description provides a set of action history, where each step corresponds to an action performed on that GUI page. The brief task is a concise task description that aligns with this GUI flow. 
  }
  \label{figure5-1}
\end{figure*}

\section{Related Work}
\textbf{Mobile AI Agent.}
Many recent studies have proposed mobile AI agents for device control. Appagent \cite{yang2023appagent} and Mobileagent \cite{ding2024mobileagent}  use prompt engineering methods and rely on existing closed-source models (e.g., GPT-4V, GPT-4o) to achieve mobile control. Other studies like CogAgent \cite{hong2023cogagent} and Agenttuning \cite{zeng2023agenttuning} use data-driven methods to fine-tune the open-source VLMs.
In order to improve the device control abilities of mobile AI agents, large-scale datasets with diverse scenarios and accurate annotations are needed.

Rico \cite{deka2017rico} and AITW \cite{rawles2023android}  are two publicly available large-scale GUI datasets, containing 72,219 single-step GUI tasks and 715,142 multi-step page navigation tasks, respectively. They are widely used in multiple GUI modeling works \cite{wang2021screen2words,li2021screen2vec,hsiao2022screenqa}.
However, they are constructed by combining crowdsourcing workers and automated annotation, which also leads to noise and wrong labels. 
To this end, a series of published works clean and filter these datasets to improve their quality.
Enrico \cite{leiva2020enrico}, UI-Bert \cite{bai2021uibert}, Vins \cite{bunian2021vins} extended the Rico dataset and proposed new GUI tasks such as UI layout classification and UI element retrieval. AutoUI \cite{zhan2023you} further filtered the GooglePlay tasks in the AitW dataset. However, these datasets mainly contain screenshots and OCR text, lacking GUI raw data such as xml documents, which limits the agent to further obtain mobile environment information. 
MobileVLM \cite{wu2024mobilevlm} and ScreenAI \cite{baechler2024screenai} build large-scale pre-training datasets to enhance the agent’s UI understanding ability. Here, Mobile3M \cite{wu2024mobilevlm} proposed by MobileVLM uses a breadth-first approach to explore the GUI pages of each APP and forms these pages into a graph structure. This allows us to obtain environmental information about different paths and operations, for example, the next pages after performing different actions in a page, and how many paths there are from one page to another.
However, as a pre-training dataset, MobileVLM lacks fine-tuning tasks. Most of their tasks are UI understanding and single-step action generation. A small number of multi-step tasks can be summarized as navigate to a page with a certain button.

Therefore, based on Mobile3M, we extracted available GUI flows and annotated a page navigation dataset. Using the graph structure environment, we further extracted page reaching and page operation datasets, and scored and paired preference datasets for RL learning. We used the three page datasets for one-stage SFT fine-tuning of ReachAgent and the preference datasets for two-stage RL optimization.

\textbf{Reinforcement Learning (RL).}
RL-based fine-tuning methods have shown great potential in improving the generation quality of LLMs \cite{liu2023rltf, shen2023pangu,yuan2023rrhf}. For mobile device control, some agents use RL methods to further optimize fine-tuned models. 

DigRL \cite{bai2024digirl} uses Gemini 1.5 Pro \cite{team2024gemini} as an automatic evaluator to assign rewards to each GUI flow generated by the agent, and updates the model with the annotated GUI flows through online RL. DistRL \cite{wang2024distrl} uses multiple workers to collect interaction data and then sends them to a central learner for learning, focusing more on distributed frameworks. These works are limited by data collection efficiency and labor-intensive manual annotation, and therefore cannot be well expanded. 

Our work combines agent generation and graph path sampling to obtain a large number of GUI flows, uses page reaching and operation subtasks to set reward rules, and builds preference data for each step of the GUI flow.

\section{Dataset Construction}
\subsection{Data Definition} \label{section3.1}
\textbf{GUI flow:}  Given a user's instruction, an agent will conduct multiple rounds of interactions with the mobile device to complete the task and record the process as GUI flow. Specifically, the GUI flow is a chain structure with the GUI page as the head node, the agent's action as the edge, and the resulting GUI page after executing the action as the tail node. Figure \ref{figure1-2} shows a GUI flow. It is defined as a 9-step chain sequence, which includes a complete set of 9 GUI Pages and 9 actions.


\textbf{GUI page:} Our algorithm uses an Android emulator to run the apps and display the GUI pages. A typical GUI page consists of a screenshot and an XML document containing multiple elements such as buttons, text boxes, icons, and images. These elements have unique identifiers, such as text, bounds, and resource IDs, used to interact with them programmatically. Figure \ref{figure4-1} shows a GUI page example.

\textbf{Action:} Our action set includes four types of actions: Click, Scroll, Input, and Complete. Figure \ref{figure5-1} summarizes the parameters and examples of these actions. When the agent interacts with the UI page, such as clicking a button or entering text, Appium\footnote{\url{https://appium.io/docs/en/latest/}} sends an action command to the Android emulator, which then performs the corresponding action. The changes of the GUI page are then captured as a new screenshot and XML document, which are used as the environment for the next step.

\subsection{Page Navigation Dataset} \label{section3.2}

Our page navigation dataset is built using Mobile3M's GUI flows.
Mobile3M \cite{wu2024mobilevlm}  uses random walks to explore the page jump paths of 49 apps and combines them into a graph format. 
We sample GUI flows from the graph and generate corresponding tasks to build the page navigation dataset. The construction pipeline is as follows:

1. GUI flow extraction: Select GUI flows with path lengths of 3-10 steps from Mobile3M.

2. Image Caption: Generate image captions for each GUI page in the flow with InternVL\cite{chen2024internvl}.

3. Task Generation: Use GPT-4V to generate a step-by-step description and a brief task for the GUI flow based on the captions of each page and the actions between pages, as shown in Figure \ref{figure5-1}. The prompt for task generation is in Appendix \ref{prompt}.

4. Data cleaning:  Filter out low-quality GUI flows with duplicate tasks or invalid GUI pages. 

After the above steps, we have a total of 53,832 GUI flows and  259,742 action steps, approximately 4.8 steps per task. Each task includes a brief task, GUI pages, and corresponding actions, along with a step-by-step description of the task.

\subsection{Page Reaching and Operation Dataset}\label{section3.3}

To improve the model's ability to reach and operate mobile GUI pages, we break down the GUI flow into multiple subtasks. There are two types of subtasks described as follows:

\noindent
\textbf{Page reaching task.} It takes an instruction for reaching a specific page as input and a GUI flow from the app homepage to that page as output. Here, each GUI page is named in three ways. 

1. If the corresponding step in the step-by-step description mentions the name of a page, we use that name directly, such as "search results page."

2. We extract the element name or text input from the action as the page name, such as "white" or "add to cart". 

3. If the name is invalid or too common, we use GPT-4 to summarize the page name based on the GUI page's caption.

We use page names and templates to construct page reaching tasks. The pre-defined task templates are shown in Appendix \ref{prompt}.

\noindent
\textbf{Page operation task.} It requires the agent to complete a specific task on a page, which requires the model to have two abilities: navigate to a page that can complete the task, and then complete the specific task. There are two extraction methods for this task:

1. In the step-by-step description, if a step involves a scroll or input action, it is usually a page operation task. We extract the sub-flow from the GUI flow to pair with this step's description. 

2. In the GUI flow, if an action has similar pages before and after execution, it is usually a page operation task. We use GPT-4V to construct a task for the sub-flow that ends with this action. 
Similar to \citet{wu2024mobilevlm}, the definition of similar pages is based on page similarity and the number of co-occurring elements. 

Eventually, we have a page reaching dataset with 67,920 GUI flows and 374,834 action steps, a page operation dataset with 76,252 GUI flows and 368,942 action steps. See Appendix \ref{construct} for details.

\begin{figure*}[!t]
  \centering
  \includegraphics[width=0.96 \textwidth]{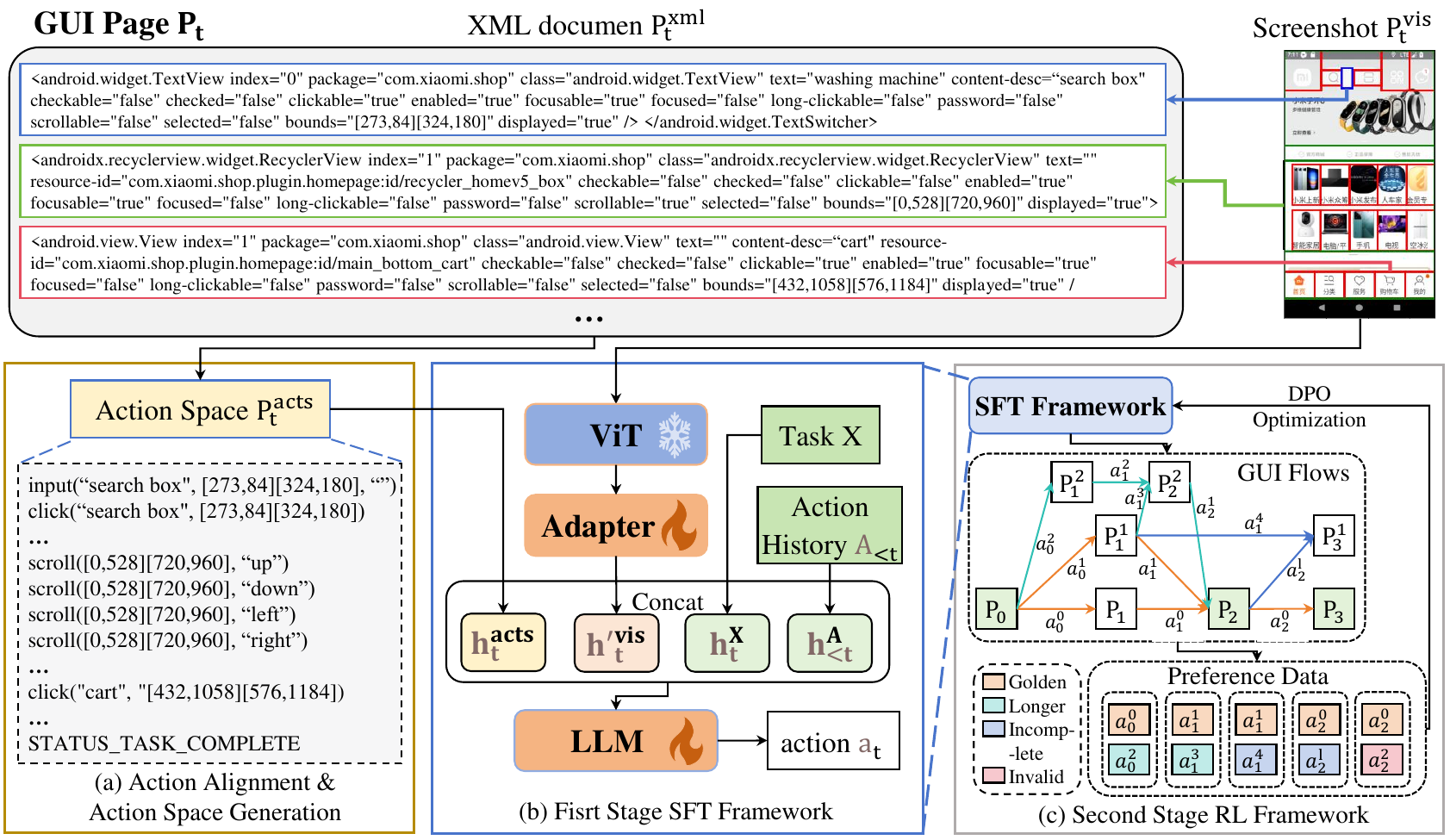}
  \caption{The overview of our proposed ReachAgent. (a) Extracting action space from XML document. (b) In the first stage, the framework generates a GUI flow through multiple interaction steps with the GUI page.  (c) In the second stage, it uses the reward function to construct preference data to further reinforce the SFT framework.
  }
  \label{figure4-1}
\end{figure*}

\section{Models}

\subsection{Action Alignment Mechanism \& Action Space Generation} \label{section4.0}
The mobile control task can be represented as a multi-round interaction process, which generates an action sequence $\overrightarrow{\text{A}}=\{a_1, a_2,..., a_t\}$ based on a given user instruction X and GUI pages $\overrightarrow{\text{P}}=\{{\text P}_1, {\text P}_2,..., {\text P}_t\}$ in the mobile environment. 
This multi-round interaction process forms a GUI flow.

\textbf{Action Alignment Mechanism.} As shown in Figure \ref{figure4-1}, the GUI page $\text{P}_t$ contains two forms: screenshot $\text{P}^{\rm vis}_t$ and XML document $\text{P}^{\rm xml}_t$. The screenshot shows multiple elements of the GUI page, while the XML document provides text descriptions and attributes for these elements.  While users can touch any point on the screen, their goal is to interact with an element on the page. To narrow down the range of candidate actions, we propose an action alignment mechanism. That is, we suggest that the agent only clicks on the center point of an element or scrolls along its central axis. For a candidate box [273,84][324,180], we can click or enter at position (298, 132). For a candidate box [0,528][720,960], we can scroll left from (360, 744) to (180, 744). 

\textbf{Action Space Generation.} With the action alignment mechanism, we can extract all the candidate actions in the GUI page as the candidate action space $\text{P}_t^{\rm acts}$. Each element in the screenshot corresponds to an element description in the XML document. We extract the element's bounding box and its attributes, such as whether it is clickable, scrollable, and inputtable, from the description and form an action. 
As shown in Figure \ref{figure4-1}, the "search box" element can be formed into an input action and a click action. The "RecyclerView" element can be formed into scrolling actions in 4 directions.

\subsection{First Stage SFT Framework}\label{section4.1}
Given a task X, in the t-th step, the screenshot $\text{P}^{\rm vis}_t$ is encoded into the image representations $\mathbf{h}^{\rm vis}_t$ by a frozen visual encoder as follows: 
\begin{equation}\label{equ1}
\setlength\abovedisplayskip{5pt}
\mathbf{h}^{\rm vis}_t=\mathbf{ViT}({\text{P}}^{\rm vis}_t).
\end{equation}
Here, visual encoder $\mathbf{ViT}$ is ViT-bigG \cite{dosovitskiy2020image}. 
The candidate action space $\text{P}_t^{\rm acts}$ is extracted from $\text{P}_t^{\rm xml}$ and encoded into the text representation $\mathbf{h}^{\rm acts}_t$ as follows:
\begin{equation}\label{equ2}
\mathbf{h}^{\rm acts}_t=\mathbf{E}(\text{ActionSpace}({\text{P}}^{\rm xml}_t)).
\end{equation}
Here, $\mathbf{E}(\cdot )$ is a wording embedding layer. The candidate action space $\mathrm{ActionSpace}({\text{P}}^{\rm xml}_t)$ is a list of all possible actions that can be used to interact with the page, as described in Section \ref{section4.0}. 


Similarly, instruction X and action history $\overrightarrow{\text{A}}_{<t}$ $=\{a_1,...,a_{t-1}\}$ are also encoded as $\mathbf{h}^{\rm X}_t,$ and $\mathbf{h}^{\rm A}_{<t}$. ReachAgent uses a position-aware visual language adapter to align visual representations with text representations, and generate the the action $a_t$:
\begin{equation}
\begin{split}
\mathbf{h}^{\rm X}_t, \mathbf{h}^{\rm A}_{<t} &=\mathbf{E}(\text{X}), \mathbf{E}(\overrightarrow{\text{A}}_{<t}), \\
\mathbf{h'}^{\rm vis}_t = & \text{Adapter}(\mathbf{h}^{\rm vis}_t) , \\
{\rm P}(a_t)=  & \text{ReachAgent}(\mathbf{h}^{\rm X}_t , 
 \mathbf{h'}^{\rm vis}_t, \mathbf{h}^{\rm acts}_t , \mathbf{h}^{\rm A}_{<t}).
\end{split}
\end{equation}

After generating the action $a_t$, we execute $a_t$ in the mobile environment to update the GUI page ${\text P}_t$ to ${\text P}_{t+1}$. Therefore, in the next step, ReachAgent has instruction X, screenshot ${\text{P}}^{\rm vis}_{t+1}$, candidate action space ${\text{P}}^{\rm acts}_{t+1}$ and action history $\overrightarrow{\text{A}}_{<t+1}$ as inputs for action $a_{t+1}$ generation. 
ReachAgent needs to go through multiple rounds of action generation and execution loops until it finally generates the action "STATUS\_TASK\_COMPLETE" to end the mobile-control process, or until it exceeds the maximum number of interaction steps.

In the first stage, ReachAgent fine-tuned its ability to reach pages and perform actions on them by using all three datasets. 
The cross-entropy loss is defined as:
\begin{equation}
{\mathcal{L}}= \sum_{t} \log \mathbf{P}(a_t| {\text{X}},{\text{P}}^{\rm vis}_{t}, {\text{P}}^{\rm acts}_{t}, {\text{A}}_{<t}).
\end{equation}

\subsection{Second Stage RL Framework}\label{section4.2}
\textbf{Preference data construction.} To further reinforce ReachAgent, we constructed a preference dataset D = \{$(\text{X},\text{P}_t,a_t^w,a_t^l)$\}. Here, $a_t^w$ is the chosen action at step t, and $a_t^l$ is the rejected action. Our construction approach is based on two key principles: 1. The GUI flow should reach the required page and complete the required operation in the instruction; 2. The GUI flow should be as short as possible.

Given an instruction X and the current page P$_t$, we can divide all potential actions into four levels:

\textbf{Golden:} The action is in the golden GUI flow, or in a GUI flow that can complete the instruction and has the same length as the golden flow.

\textbf{Longer:} The action is in a GUI flow longer than golden, but can still complete instruction X.

\textbf{Incomplete:} The action is not in a GUI flow that can complete the instruction.

\textbf{Invalid:} The action cannot be executed or is not in the action space of the current page.

Naturally, for these four levels, we expect their reward scores to be:
\begin{equation}
\setlength\abovedisplayskip{5pt}
\resizebox{.89\hsize}{!}{\text{R(Golden)>R(Longer)>R(Incomplete)>R(Invalid)}}
\end{equation}

Given a task, "add product A to the shopping cart", the subtasks it needs to complete are simplified into Reach(P$_2$): Go to the "Product A" page, and Operate(P$_2$, "add to cart"): Add Product A to the cart on the product page.
Figure \ref{figure4-1} shows several GUI flows for this task. "P$_0$, P$_1$, P$_2$, P$_3$" and "P$_0$, P$_1^1$, P$_2$, P$_3$" are the 4-step golden GUI flows. "P$_0$, P$_1^2$, P$_2^2$, P$_2$, P$_3$" takes 5 steps to complete the task and marked as Longer. "P$_0$, P$_1^1$, P$_3^1$" doesn't reach the P$_2$ page and marked as Incomplete.

Based on the 4-level reward function, we identify the chosen and rejected actions for the shared history flow. Start with GUI page $\text{P}_0$, $a_0^0$ and $a_0^1$ are in the Golden flow, while $a_0^2$ leads to a Longer flow. Therefore, R($a_0^0|\text{P}_0$)>R($a_0^2|\text{P}_0$) and  ($\text{X},\text{P}_0,a_0^0,a_0^2$) is a preference data. 
Start with page $\text{P}_1^1$, $a_1^1$, $a_1^3$, and $a_1^4$ lead to a Golden, Longer, and Incomplete flow respectively. The preference data could be ($\text{X},\text{P}_1^1,a_1^1,a_1^3$) and ($\text{X},\text{P}_1^1,a_1^1,a_1^4$).

To collect preference data, we use the SFT agent to regenerate the training split of the page navigation dataset. 
If the reward of an action is less than the golden action, we use it to construct the preference data. Otherwise, we randomly select an action with a lower reward in the action space of each page to pair with the golden action. 


\begin{table*}[!tb]
    \centering
\resizebox{0.85\textwidth}{!}{
\begin{tabular}{ll|cc|cc|c}
\hline\hline
        \multirow{2}{*}{Model}  & \multirow{2}{*}{Method}   &  \multicolumn{2}{c|}{Step-level Acc} &  \multicolumn{2}{c|}{Task-level Acc} & Task Success  \\  \cline{3-7} 
        & &   IoU & Text & IoU  & Text   \\ \hline 
        GPT-4o \cite{openai2023gpt4} & FewShot &19.44\% & 17.06\% & - & - & - \\ 
        Qwen-VL  \cite{bai2023qwen} & FewShot  & 0.05\% & 1.49\% & - & - & - \\ 
        MobileVLM$\rm _{unified}$ \cite{wu2024mobilevlm}  & FewShot & 2.43\% & 4.91\% & - & - & - \\ 
        MobileVLM$\rm _{seperate}$ \cite{wu2024mobilevlm}  & FewShot  & 1.75\% & 10.60\% & - & - & - \\  \hline
        Qwen-VL \cite{bai2023qwen}  & SFT & 73.38\% & 72.14\% & 30.13\% & 26.22\% & 35.77\% \\ 
        Auto-UI$\rm _{unified}$ \cite{zhan2023you} & SFT  & 73.26\% & 70.88\% & 29.60\% & 22.20\%   & 33.40\% \\
        MobileVLM \cite{wu2024mobilevlm}   & SFT & 76.20\% & 74.08\% & 34.03\% & 28.43\% & 39.78\% \\ \hline 
        \textbf{ReachAgent-stage 1} &SFT & \textbf{83.34\%} & 81.47\% & 37.82\% & 31.31\% & 44.85\%  \\ 
        \textbf{ReachAgent-stage 2} &SFT+RL  & 83.32\% & \textbf{81.77\%} & \textbf{38.75\%} & \textbf{33.06\%} & \textbf{46.37\%} \\ 
        \hline\hline
\end{tabular}
    }
    \caption{Main Result(\%) on MobileReach. SFT baselines are fine-tuned for 2 epochs on page navigation split. 
    }
    \label{result_1}
\end{table*}

\textbf{DPO Optimization.}  Direct Policy Optimization (DPO) \cite{rafailov2024direct} does not require an explicit reward score, but only requires a preference for paired data, which is more suitable for our 4-level reward function.
Therefore, we adopt DPO to optimize ReachAgent.
The DPO loss is defined as:
\begin{equation}
\setlength\belowdisplayskip{5pt}
\begin{split}
{\mathcal{L}}_{\rm DPO}(\pi_{\theta}; \pi_{\rm SFT})  = &- \mathbb{E}_{(\text{X},\text{P}_t,a_t^w,a_t^l)\sim D} [\log_{} \sigma \\
(\beta \log_{} \frac{\pi_{\theta}(a_t^w|\text{X},\text{P}_t)}{ \pi_{\rm SFT}(a_t^w|\text{X},\text{P}_t)}  &- \beta \log_{} \frac{\pi_{\theta}(a_t^l|\text{X},\text{P}_t)}{ \pi_{\rm SFT}(a_t^l|\text{X},\text{P}_t)}) ].\\
\end{split}
\end{equation}
Here, $\pi_{\theta}$ is the DPO agent to learn and $\pi_{\text{SFT}}$ is the first stage SFT agent. $\theta$ is the sigmoid function and $\beta$ is a hyperparameter that controls the deviation from $\pi_{\rm SFT}$.



\begin{table}[!tb]
    \centering
\resizebox{1\columnwidth}{!}{
    \begin{tabular}{lccccc}
        \hline
        \multirow{2}{*}{Dataset} & \multicolumn{2}{c}{Train} & \multicolumn{2}{c}{Test}   \\ \cline{2-3} \cline{4-5}
         & Chain & Step & Chain & Step  \\  \hline
        Page Navigation & 53,832 & 259,742 & 2,689 & 12,922  \\ 
        Page Reaching & 67,920 & 374,834 & 3,385 & 16,253  \\ 
        Page Operation & 76,252 & 368,942 & 3,798 & 18,338 \\
         \hline
    \end{tabular}}
    \caption{The statistics of MobileReach dataset.}
    \label{table5_1}
\end{table}

\section{Experiment}
\subsection{Datasets and Benchmarks}
The MobileReach dataset contains 3 splits: page navigation, page reaching, and page operation.  We used the GUI graph from Mobile3M to build the MobileReach dataset. 
The detailed information of the MobileReach dataset is shown in Table \ref{table5_1}. For preference data, 48,013 rejected actions are generated by the agent, and 211,729 rejected actions are sampled from the action space.

We used the page navigation split of the MobileReach dataset and the Auto-UI dataset \cite{zhan2023you} for testing. For the Auto-UI dataset, we follow the official split and method for finetuning. For the MobileReach dataset, we randomly selected 5\% of the GUI flows as a test set before data construction. Therefore, there was no overlap between the training set and the test set.

\subsection{Evaluation Metrics} \label{section5.2}
We use two objective metrics to evaluate the position and text of the generated action. IoU Acc evaluates the bounding box accuracy. Text Acc evaluates the text accuracy. 
Step-level accuracy evaluates whether each action in a GUI flow is correct, while task-level accuracy evaluates whether all actions in an entire GUI flow are correct. See more metric details in Appendix \ref{baselines}.


\subsection{Implementation Details and Baselines}
ReachAgent chose MobileVLM as the backbone model. 
We use 8 80GB Nvidia A100 GPUs for fine-tuning. The learning rate is 1e-5, and the agent's max length is 4,096. 
For the MobileReach dataset, in the first stage, ReachAgent was trained for 2 epochs on all three splits. In the second stage, it was trained for 2 epochs on 259,742 preference data. 
For the Auto-UI dataset, ReachAgent is fine-tuned for 1 epoch on its training data using the first-stage framework.
For specific information on parameters and baselines, refer to Appendix \ref{baselines}.

\begin{table*}[!tb]
    \centering
\resizebox{1\textwidth}{!}{
\begin{tabular}{lcccc|cc}
\hline\hline
        \multirow{2}{*}{Model}   &  \multicolumn{2}{c}{Reach SubTask} &  \multicolumn{2}{c|}{Operate SubTask}  & \multicolumn{2}{c}{All SubTask}  \\ \cline{2-7} 
         &  IoU Acc & Text Acc & IoU Acc & Text Acc & IoU Acc & Text Acc  \\ \hline 
MobileVLM$\rm _{SFT}$ & 51.44\% & 53.87\% & 56.62\% & 37.89\% & 53.19\% & 48.48\% \\
 \quad + action alignment & 53.90\% & 54.91\% & 53.97\% & 39.39\% & 53.92\% & 49.82\% \\
 \quad + action alignment \& page reaching & 55.79\% & 56.83\% & 59.57\% & 43.39\% & 57.03\% & 52.42\% \\
 \quad + action alignment \& page
operation & 57.36\% & 58.24\% & 60.93\% & 46.42\% & 58.53\% & 54.37\%  \\
 \quad + action alignment \& page reaching \& page
operation & 57.84\%  & 58.94\% & 59.96\% & 43.33\% & 58.53\% & 53.82\% \\  \hline
ReachAgent & \textbf{59.91\%} & \textbf{60.57\%} & \textbf{62.35\%} & \textbf{47.45\%} & \textbf{60.71\%} &  \textbf{56.27\%} \\
        \hline\hline
\end{tabular}
    }
    \caption{The task level accuracy (\%) of different ablations for completing subtasks. 
    }
    \label{result_3}
\end{table*}

\begin{table}[!tb]
    \centering
\resizebox{1\columnwidth}{!}{
\begin{tabular}{l|ccc|c}
\hline\hline
        Model  & General & Install & GoogleApps & Overall  \\ \hline
        GPT-4V  & 43.01\% & 46.14\% & 49.18\% & 48.91\%  \\ 
        Qwen-VL Max  & 46.22\% & 50.30\% & 49.16\% & 49.21\%  \\ 
        GPT-4o  & 47.06\% & 49.12\% & 52.30\% & 52.04\%  \\ \hline
        Llama 2+plan  & 53.77\% & 69.10\% & 61.19\% & 61.57\%  \\ 
        MobileAgent  & 55.80\% & 74.98\% & 63.95\% & 64.51\%  \\ 
        Auto-UI$\rm _{unified}$  & 68.24\% & 76.89\% & 71.37\% & 71.71\%  \\ 
        CoCo-LLaVA  & 58.93\% & 72.41\% & 70.81\% & 70.72\%  \\ 
        CogAgent  & 65.38\% & 78.86\% & \textbf{74.95\%} & 75.05\%  \\ 
        MobileVLM  & 69.58\% & 79.87\% & 74.72\% & 74.99\%  \\ \hline
        \textbf{ReachAgent}  & \textbf{70.27\%} & \textbf{80.76\%} & 74.94\% & \textbf{75.27\%} \\ \hline
        \hline\hline
\end{tabular}
    }
    \caption{Main Result(\%) on Auto-UI dataset. The first column shows few-shot agents. The second column shows the SFT agents. }
    \label{result_1_1}
\end{table}

\begin{table}[!tb]
    \centering
\resizebox{1\columnwidth}{!}{
\begin{tabular}{lcccc}
\hline\hline
        \multirow{2}{*}{Model}   &  \multicolumn{2}{c}{Step-level} &  \multicolumn{2}{c}{Task-level}   \\ \cline{2-5} 
         &  IoU & Text & IoU & Text  \\ \hline 
        MobileVLM$\rm _{SFT}$ & 76.20\% & 74.08\% & 34.03\% & 28.43\% \\
 \;+ Al & 81.55\% & 79.86\% & 35.37\% & 30.01\% \\
 \;+ Al \& Re & 82.34\% & 80.49\% & 36.37\% & 29.64\% \\
 \;+ Al \& Op  & 82.54\% & 81.22\% & 37.19\% & 32.87\% \\
 \;+ Al \& Re \& Op & \textbf{83.34\%} & 81.47\% & 37.82\% & 31.31\% \\ \hline
ReachAgent & 83.32\% & \textbf{81.77\%} & \textbf{38.75\%} & \textbf{33.06\%}  \\
        \hline\hline
\end{tabular}
    }
    \caption{Ablation study on the action alignment mechanism (Al), page reaching subtask (Re), page operation subtask (Op), and reinforcement learning.}
    \label{result_2}
\end{table}

\subsection{Main results}
The main experimental results are presented in Table \ref{result_1} and \ref{result_1_1}. We can observe that: 

$\bullet$ For MobileReach dataset, ReachAgent improves the IoU Accuracy, and Text Accuracy by \textbf{+7.12\%}, \textbf{+7.69\%} on step-level and \textbf{+4.72\%} and \textbf{+4.63\%} on task-level compared to fine-tuned MobileVLM. It also raises \textbf{+6.59\%} on Task Success Rate. This proves that ReachAgent is not only good at generating actions at each step, but also provides a more efficient GUI flow to successfully complete the task.

$\bullet$ At both step-level and task-level, ReachAgent-stage1 has a significantly higher IoU and Text accuracy than other SFT baselines. 
We attribute this to our action alignment mechanism and the introduction of page reaching and page operation subtasks. ReachAgent achieves better page navigation abilities by learning how to go to specific pages and complete specific operations.

$\bullet$ Compared to ReachAgent-stage1, ReachAgent significantly outperforms it at the task level and is competitive with it at the step level. This can be attributed to our reinforcement learning mechanism, which encourages the agent to generate GUI flows that can complete the task. 

$\bullet$ For the Auto-UI dataset, ReachAgent outperformed the SOTA baselines in all tasks. 
Note that ReachAgent achieves this result without building subtask data on Auto-UI. This proves the generalizability of the agent's page reaching and page operation abilities.



\begin{figure*}[!t]
  \centering
  \includegraphics[width=0.8\textwidth]{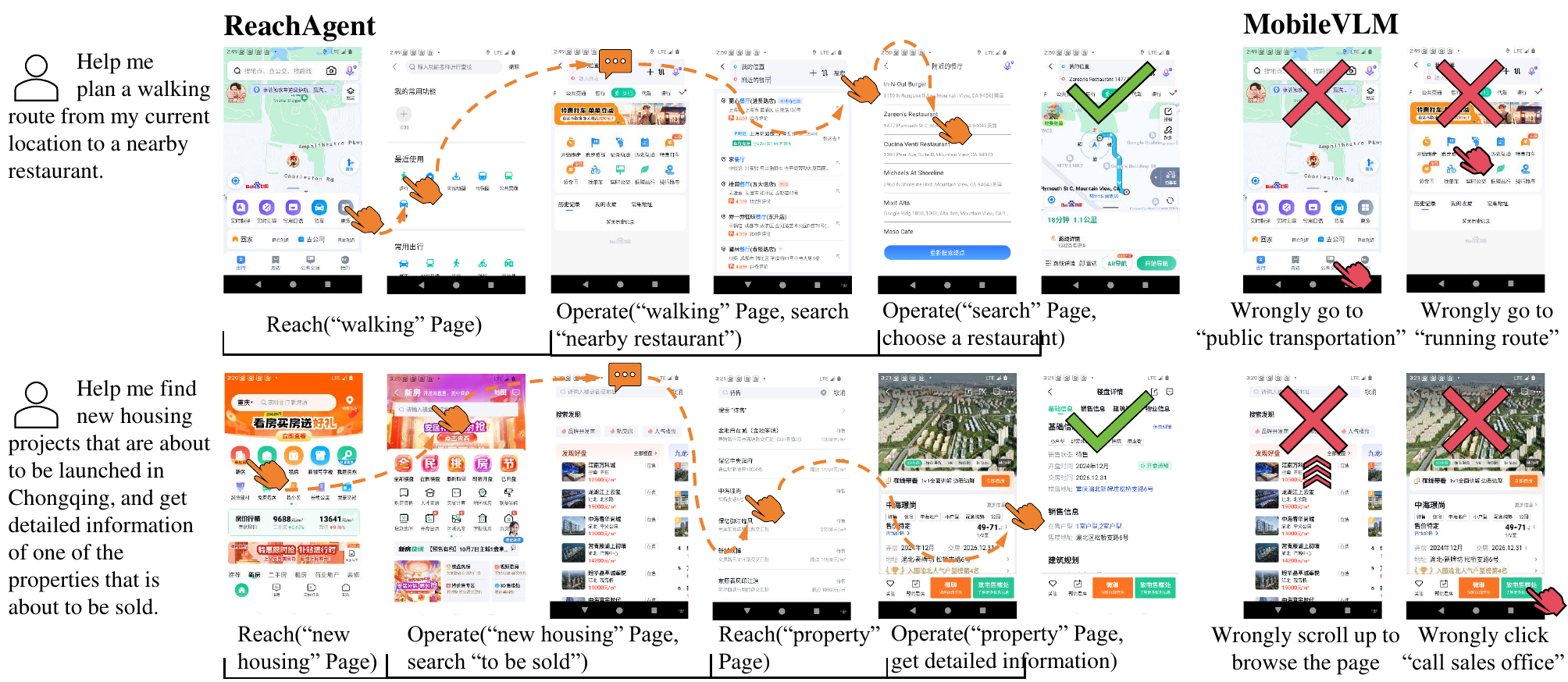}
  \caption{
  Two cases of generated GUI flow by ReachAgent and MobileVLM.
  }
  \label{result_4}
\end{figure*}

\subsection{Ablation Study}
Table \ref{result_2} shows several ablation experiment results.

\noindent
\textbf{Effects of Action Alignment Mechanism:}
Through the action alignment mechanism, the performance of the agent (+Al) is improved by 5.55\% and 1.46\% on average at the step level and task level. This is because the action alignment mechanism helps to narrow down the range of possible actions that the agent can take, reducing the generation difficulty.

\noindent\textbf{Effects of Page Reaching and Page Operation Subasks:}
Adding the page reaching subtask and the page operation subtask both improve the agent's performance.
The accuracy of the "+ Al \& Re \& Op" increases by an average of 1.7\% and 1.87\% at the step level and task level compared to "+ Al". 
These subtasks enhance the agent's understanding of the task goal and the order of actions, making the agent generate actions that are more aligned with the overall goal of the task. This leads to better performance at both the step and task levels.

\noindent\textbf{Effects of Reinforcement Learning:} ReachAgent further adds RL to "+ Al \& Re \& Op". It mainly improves the task-level accuracy. This is because our reward focuses more on whether the agent can generate a short GUI flow that can complete all subtasks, rather than whether the action of each step is exactly the same as the golden answer.

\noindent\textbf{Ability to complete subtasks:} We also compared the ability of different ablations to complete the Reach and Operate subtasks. As shown in Table \ref{result_3}, ReachAgent improves the overall IoU accuracy by 1.9\% and the Text accuracy by 2.18\%, outperforming the agent without RL in all subtasks. This shows that the RL mechanism improves the agent's ability to complete various subtasks. In addition, adding the page operation subtask contributes more to the agent's performance than adding the page reaching subtask. This is because the page operation subtask not only improves the agent's ability to operate pages, but also requires the agent to be able to reach the specified page. 

\subsection{Case Study}

Figure \ref{result_4} shows two generation cases.
MobileVLM tends to greedily choose task-related actions at each step, leading to local optima. In the first case, it goes directly to the "Public Transportation" page on the homepage, or clicks on "Running Routes" before finding nearby restaurants. In the second case, when the task is to search for a property that is about to be sold, MobileVLM keeps scrolling down, trying to browse directly to find the required property, and gets stuck in an endless loop.
In contrast, ReachAgent improves the ability to reach and operate on task-related pages, and further completes the task based on these subtasks.

\section{Conclusion}
In this work, we propose ReachAgent, a two-stage framework that leverages page reaching and page operation tasks to enhance the agent's subtask completion abilities.
Besides, we use a 4-level reward function to collect preference GUI flows, and further enhance the agent's overall task completion ability with reinforcement learning.
In addition, we construct a mobile control dataset called MobileReach, which contains 3 categories of tasks: page navigation, page reaching, and page manipulation.
Experimental results show that ReachAgent significantly improves IoU and Text accuracy by \textbf{7.12\% and 7.69\%} at the step level and \textbf{4.72\% and 4.63\%} at the task level, respectively.

Our ReachAgent shows strong task completion abilities by addressing the challenge of mobile AI agents focusing more on single-step action accuracy rather than completing the entire task flow. We hope that MobileReach can serve as a useful resource for breaking down tasks, solving page reach and operation subtasks, and provide assistance for future research.

\clearpage

\section*{Limitations}
Our training data is built on a graph consisting of GUI flows of 49 commonly used apps. Since the method of exploring the GUI graph is random walk, this may not cover all the functions of the app. In addition, we select GUI flows by random sampling, which may result in many invalid GUI flows that do not have corresponding tasks.

\section*{Ethics Statement}
This paper was conducted in accordance with the ACM Code of Ethics. Our MobileReach dataset is constructed using publicly available platforms and data sources, ensuring that there are no privacy issues or violations. All data used in our research was obtained following legal and ethical standards, and we do not collect any personally identifiable information. 

\bibliography{acl_latex}

\clearpage

\appendix

\section{Experiment Settings}\label{baselines}
\subsection{Baselines}
ReachAgent was compared to four other baselines as follows:
GPT-4o, Qwen-VL, Auto-UI, MobileVLM. 
\begin{itemize}
    \item GPT-4o, GPT-4V \cite{openai2023gpt4} are most advanced VLMs currently available.
    \item Qwen-VL, Qwen-VL-Max \cite{bai2023qwen} are large-scale visual language models designed to perceive and understand text and images. They have demonstrated significant performance in tasks such as image captioning, question answering, visual or document visual question answering, and localization, and have been applied as the base model for multiple mobile AI agents.
    \item Auto-UI \cite{zhan2023you} is derived from AITW \cite{rawles2023android}, which uses a chain-of-action composed of a series of action histories and future action plans to improve the agent's action prediction ability.
    \item MobileVLM \cite{wu2024mobilevlm} builds a large-scale Mobile3M dataset in the graph structure and uses multiple pre-training tasks to enhance the agent's UI understanding ability.
    \item MobileAgent \cite{ding2024mobileagent} builds a GUI agent based on prompt engineering based on GPT-4V.
    \item CoCo-LLaVA \cite{ma2024coco} leverages comprehensive environment perception (CEP) and conditional action prediction (CAP) to enhance the agent’s understanding of GUI pages and tasks.
    \item CogAgent \cite{hong2023cogagent} is built on CogVLM, adding pre-training tasks and supporting higher-resolution images.
\end{itemize}
For in-context learning, we provided them with several few-shot examples. For SFT learning, we use the page navigation split of the MobileReach dataset to fine-tune them for two epochs.
We maintain consistent hyperparameters across all the baselines for fair comparisons. 

\subsection{Metric}
We use two objective metrics to evaluate the model at the step-level:

\textbf{IOU Acc:} Intersection over Union \cite{cheng2021boundary} evaluates if the bounding box of the generated action intersects with the golden action. We allow a 14\% margin of error relative to the screen size.

\textbf{Text Acc:} It evaluates if the text in the generated action matches the golden one. This includes the name of the click action, the direction of the scroll action, the name and input text of the input action, and the text of the complete action. For the input text, which may have varied descriptions, we require an F1 value greater than 0.8 to be considered consistent. The remaining text must be entirely consistent. 

We use two metrics to evaluate the model at the task-level:

\textbf{Task-level Acc}: As described in Section \ref{section5.2}, task-level accuracy evaluates whether all actions in an entire GUI flow are correct. That is, all actions in the reasoning process are exactly the same as all actions in the ground truth. This is a very strict method for judging task success and has been applied in previous works.

\textbf{Task Success Rate}: This metric considers a GUI flow that completes all subtasks as correct.
In our reward scoring principle, if the model can complete all subtasks, it does not need to be exactly the same as the ground truth to be considered complete. However, since the Auto-UI dataset does not contain explored paths other than the golden answer, we only use this evaluation metric in the Mobile3M dataset.

\begin{figure*}[!t]
  \centering
  \includegraphics[width=1 \textwidth]{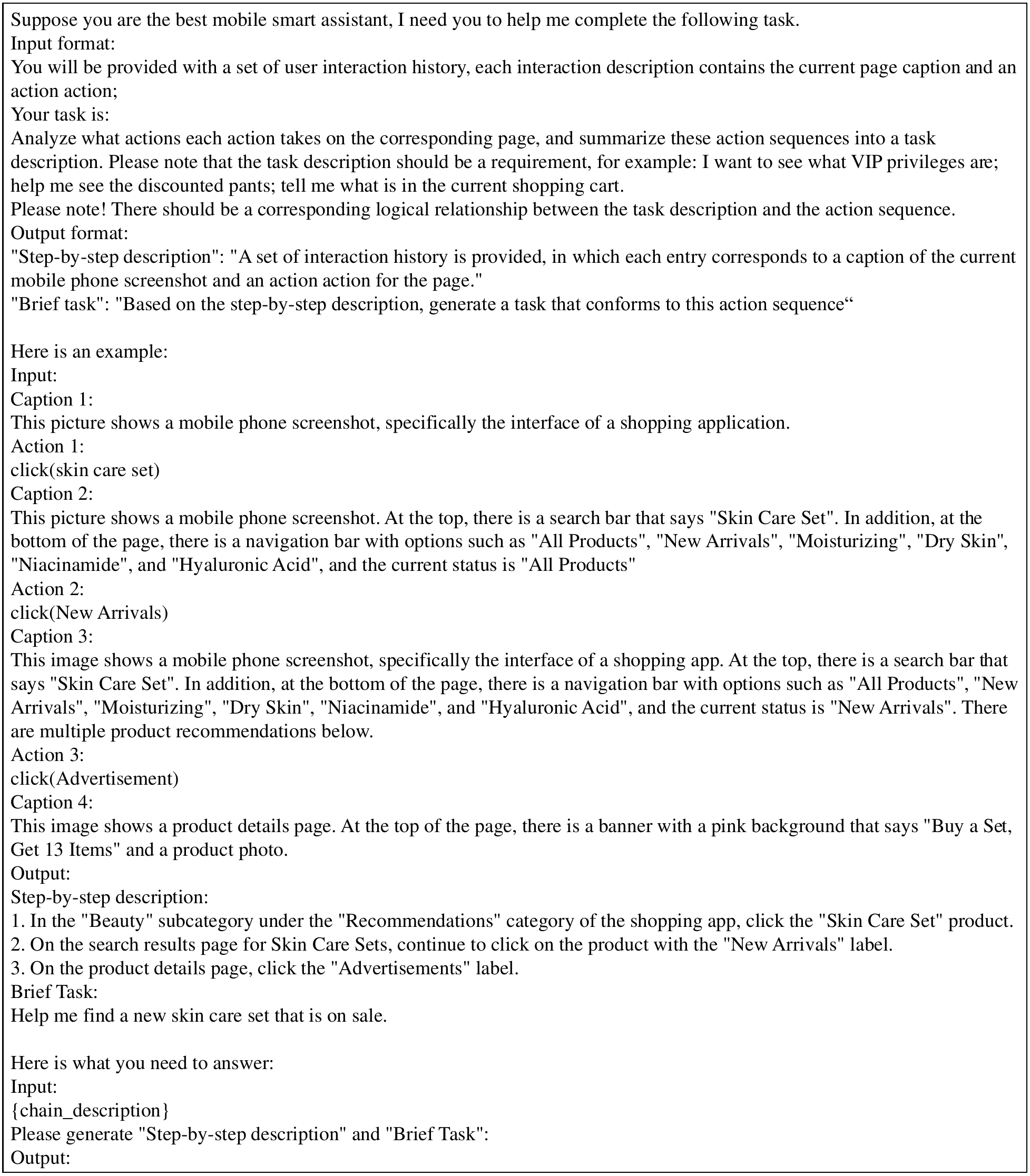}
  \caption{The prompt for step-by-step description and brief task generation.
  }
  \label{figure6-1}
\end{figure*}

\subsection{HyperParameter}
We present the hyperparameters for the first stage SFT framework and second stage RL framework in Table \ref{parameters}.
ReachAgent chose MobileVLM as the backbone model. 
We use 8 80GB Nvidia A100 GPUs for fine-tuning. The learning rate is 1e-5, and the agent's max length is 4,096. 

For the MobileReach dataset, in the first stage, the model was trained for 2 epochs on all three splits with a batch size of 4. In the second stage, it was trained for 2 epochs on 259,742 preference data with a batch size of 1. During testing, the max step is set to 15. 

For the Auto-UI dataset, we fine-tuned the MobileReach-stage1 on its training split for 1 epoch. Similar to its official method, we only used 10\% of the GoogleApps data of AITW to save the training time. All SFT baselines are also fine-tuned for 1 epoch on this split with the same parameters.  During testing, we removed WebShopping tasks because they are control tasks performed on a computer screen rather than a mobile screen.

\begin{table}[!htb]
    \centering
\resizebox{\columnwidth}{!}{
    \begin{tabular}{l|c|c}
        \hline
Hyperparameter & Stage 1  & Stage 2\\
        \hline
epoch & 2 & 2 \\
batch\_size & 4 & 1 \\
learning\_rate & 1e-5 & 1e-5\\
warmup\_ratio & 0.02 & 0.02\\
optimizer & Adam & Adam \\
max\_sequence\_length & 4096 & 4096 \\
GPUs  & 8 & 8 \\
         \hline
    \end{tabular}}
    \caption{Hyperparameters.}
    \label{parameters}
\end{table}

\section{Prompt}\label{prompt}
For the \textbf{page navigation task}, we use the few-shot prompt to guide the GPT-4 in building tasks, as shown in Figure \ref{figure6-1}. The inputs are image captions generated by Intern-VL and actions between GUI pages. The outputs are step-by-step description and brief task.

For the \textbf{page reaching task}, we use the following pre-defined task templates to build tasks. We replace the \{text\} in the template with page names.

\begin{mdframed}[linewidth=1pt,linecolor=black]
Navigate to \{text\} page.\\
Go to \{text\} page.\\
From the current page, what interactions should be performed to reach \{text\} page? \\
What actions need to be performed to reach \{text\} image? \\
Determine the actions that need to be taken to display \{text\} page. \\
Visit the \{text\} page. \\
What actions should you take to advance to the page showing \{text\}? \\
I want to go to \{text\} interface. \\
What actions will take you to \{text\} image? \\
Describe the steps that need to be taken on the current image to find \{text\} image. \\
Is the page showing \{text\}? \\
First, find \{text\} page. \\
Help me find the page with \{text\}. \\
Perform a series of actions to reach \{text\}. \\
First visit \{text\} page. \\
What actions do I need to take to find \{text\} page? \\
How do I get to \{text\} page? \\
Help me navigate to \{text\} interface. \\
Go to \{text\} interface. \\
Jump to \{text\} page. \\
Next, enter \{text\} page. \\
Visit the page showing \{text\}? \\
Find the image with \{text\}? \\
How to get to the page with \{text\}? \\
I want to go to \{text\} page. \\
Open \{text\} image? \\
Next, go to \{text\} page. \\
Need to visit \{text\}. \\
Enter \{text\} page. \\ 
Navigate to \{text\}. \\
How to get to the page with \{text\}? \\
Guide to the image with \{text\}. 
\end{mdframed}

\section{Page Reaching and Operation Dataset Construction} \label{construct}
\subsection{Subflow Filtering}
\begin{figure*}[!t]
  \centering
  \includegraphics[width=1 \textwidth]{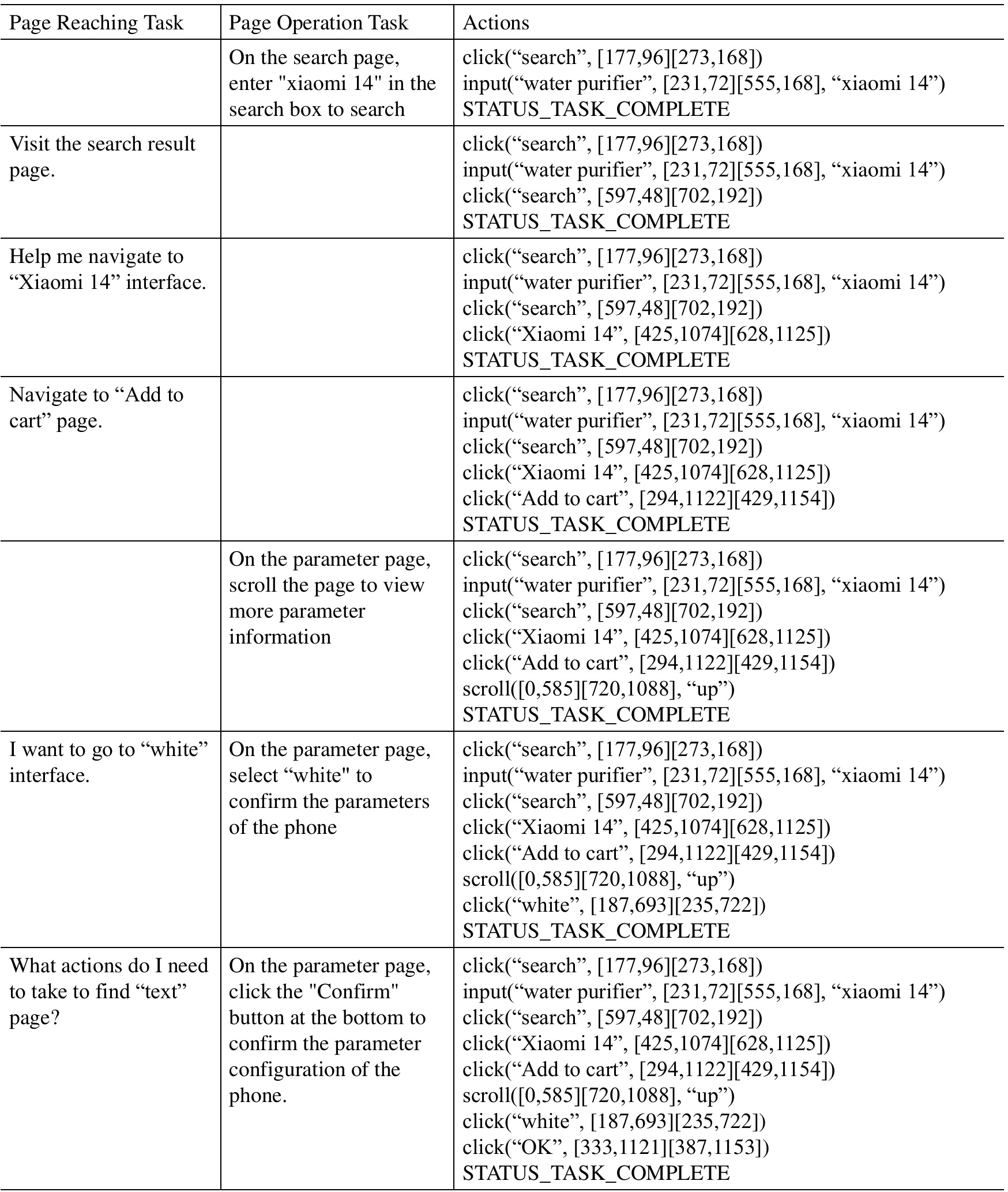}
  \caption{Page reaching subtasks and page operation subtasks extracted from the GUI flow example in Figure 3.
  }
  \label{figure7-1}
\end{figure*}

Figure \ref{figure7-1} shows the page reaching subtasks and page operation subtasks extracted from the GUI flow example in Figure 3.

Here, we provide a step-by-step data construction process from GUI flow to task and subtask generation. Take the case in Figure 3 and Figure 7 as an example:

1. First, we will observe such a GUI flow:

\begin{table}[!h]
    \centering
    \resizebox{\columnwidth}{!}{
    \begin{tabular}{|l|l|}
    \hline
         GUI Page & Action   \\ \hline
         P\_1 & a\_1, click(“search”, [177,96][273,168])   \\ \hline
         P\_2 & a\_2, input(“water purifier”, [231,72][555,168],  \\ 
         &“xiaomi 14”)   \\ \hline
         P\_3 & a\_3, click(“search”, [597,48][702,192])   \\ \hline
         P\_4 & a\_4, click(“Xiaomi 14”, [425,1074][628,1125])   \\ \hline
         P\_5 & a\_5, click(“Add to cart”, [294,1122][429,1154])   \\ \hline
         P\_6 & a\_6, scroll([0,585][720,1088], “up”)   \\ \hline
         P\_7 & a\_7, click(“white”, [187,693][235,722])   \\ \hline
         P\_8 & a\_8, click(“OK”, [333,1121][387,1153])   \\ \hline
         P\_9 & -  \\ \hline
    \end{tabular}}
\end{table}

2. Then,  we need to check the validity of this flow, generate corresponding tasks, and filter. 

\textbf{Check the validity}

$\bullet$ Whether this path of $P_1$, $P_2$, ..., $P_9$ has already appeared in the current training set.

$\bullet$ Whether all actions in the action sequence have already appeared in the current training set.

$\bullet$ Whether the action sequence contains consecutive repeated actions.

$\bullet$ Whether there is an action $a_i$ that is not included in the action space of $P_i$

\textbf{Generate Corresponding Tasks.}
We use InternVL to genenrate image captions for each GUI page in the flow and  use GPT-4V to generate a step-by-step description and a brief task as follows:

\textbf{Step-by-step Description:}

1. On the homepage of Xiaomi Mall, click the search icon to enter the search page. 
	
2. On the search page, enter "xiaomi 14" in the search box to search. 

3. On the search page, click the search icon to search. 

4. On the search results page, select the detailed information of "xiaomi 14".

5. On the detailed information page, select and click "Add to Cart" to enter the parameter page of the phone.	

6. On the parameter page, scroll the page to view more parameter information.

7. On the parameter page, select “white" to confirm the parameters of the phone.
	
8. On the parameter page, click the "Confirm" button at the bottom to confirm the parameter configuration of the phone.  
    
\textbf{Brief Task:} 

Help me find detailed information about xiaomi 14, and add a white one to the shopping cart.

Here, each step $s_i$ corresponds to a GUI Page $P_i$ and an action $a_i$ in the GUI flow.

\textbf{Filter:}

We filter out some tasks and GUI flows with low quality. For example:

A. The brief task already appears in the current training set.

B. The number of steps in the distribution description is inconsistent with the number of steps in the GUI flow.

C. The brief task is too long.

\subsection{Subtask Generation}

After obtaining the GUI flow, step-by-step description, and a brief task, we can split them into page arrival and page operation subtasks.
This splitting operation is mainly based on the step-by-step description and actions.

\textbf{Page Reaching Subtask Generation}

1. If a page is referred to by a unique name in the step-by-step description, we split out the sub-page flow leading to that page and assign a task through template to reach that page. For example:

Description:  4. On the search results page, select the detailed information of "xiaomi 14".

Task:  Visit the search result page.

GUI Flow: $P_1$ -> $a_1$ -> $P_2$ -> $a_2$ -> $P_3$ -> $a_3$ -> $P_4$

2. If the element name of a click action does not appear in the existing dataset, we name the next page of clicking on the element with that element name and assign a task to reach that page. For example:

Action:  $a_4$, click(“Xiaomi 14”, [425,1074][628, 1125]).

Task:  Help me navigate to “Xiaomi 14” interface.

GUI Flow: $P_1$ -> $a_1$ -> $P_2$ -> $a_2$ -> $P_3$ -> $a_3$ -> $P_4$ -> $a_4$ -> $P_5$

\textbf{Page Operation Subtask Generation}

1. If the action type is scroll or input, we assign an operation task as referenced in the step-by-step description, including reaching the current page and performing this action.

Action:  $a_6$, scroll([0,585][720,1088], “up”).

Task:  On the parameter page, scroll the page to view more parameter information.

GUI Flow: $P_1$ -> $a_1$ -> $P_2$ -> $a_2$ -> $P_3$ -> $a_3$ -> $P_4$ -> $a_4$ -> $P_5$ -> $a_5$ -> $P_6$ -> $a_6$ -> $P_7$

2. If an action has similar GUI pages before and after execution, we assign an operation task according to the step-by-step description,  describing the process of reaching this page to perform this action.

Similar Pages:  $P_7$, $P_8$

Task:  On the parameter page, select “white" to confirm the parameters of the phone.

GUI Flow: $P_1$ -> $a_1$ -> $P_2$ -> $a_2$ -> $P_3$ -> $a_3$ -> $P_4$ -> $a_4$ -> $P_5$ -> $a_5$ -> $P_6$ -> $a_6$ -> $P_7$ -> $a_7$ -> $P_8$



\end{document}